# Conscious Intelligent Systems

Artificial Intelligence and Consciousness – A Learning System Perspective

## Part II: Mind, Thought, Language, and Understanding

**Preface**

**Understanding**

**Communication of Understanding**

**Minds Necessitate Language**

**The Humanoid Fantasies**

## Mind, Thought, Language and Understanding

**Preface**

This is a companion paper to Conscious Intelligent Systems Part 1 by the same author (1), which discusses a possible evolutionary path for consciousness and intelligence from simple systems to human level consciousness and intelligence. Man has long been held to be a thinking animal, his thought processes have been held to be the reason for his superiority over the animals. The grand aim of AI has always been to make an entity that can think. Turing took up this very question in his paper (2) on whether machines can think.

On the more prosaic roads that real AI has been forced to follow, such grand questions have almost died down. Another major trigger for the demise has been Searle's Chinese Room (3) parody. With this rather cunning device, Searle set the cat among the pigeons and has helped induce self-doubt in the best of AI theorists. One of the major triggers towards Searle's views was language, whether syntax suffices for semantics and therefore understanding.

From our evolutionary learning system perspective, which we discuss in Part I of this discussion, we see that all these processes are tied together, the processes of consciousness, intelligence, mind, thought, and language. In a bid to show the interconnectedness of these factors, we take up the question of understanding and its communication. Similar to our treatment of the subject of consciousness based intelligent systems in Part 1, here we treat understanding from first principles.

**Understanding**

In the real world when we use the term understanding, it has two main attributes; one is the capacity to infer, the other is the capacity to recognize or discern. In computing and AI contexts the word understanding is arguably tilted more in favor of inference than perception or cognition, in normal life and in the natural kingdom the reverse is true. This is primarily because AI's aims and present status look elemental when compared to the entities of the natural world. The other reason is that AI entities find it easier to infer than cognize, which is in itself a reflection of their design sources and its aims.

For the purposes of this discussion the term understanding implies the natural version, a mix of cognition and inference. If we start from first principles, it is clear that for a rule to emerge out of a set of raw data, an inferential process has to run on it. This process could be a formal inferential process or a process that is driven by the needs of economy or efficiency.

Rules need not always rise out of intentional activity, for instance the interaction of water flowing from an open tap into a pot already full of water can create a set of rules that disallow further water entry, limit mixing and regulate overflow, many natural rules rise from interactions like these. Intentional processes however help create rules that retain their logical structure and remain stable in the long run. For our discussion a rule denotes the result of an inferential process running on raw data.

# Conscious Intelligent Systems
Artificial Intelligence and Consciousness – A Learning System Perspective
## Part II: Mind, Thought, Language, and Understanding

Once such a rule is generated, then it straddles all data that are similar to it, irrespective of whether such data is in its history or its future. This means that fresh data that are similar to the one straddled by a rule do not need further processing, the rule is simply reapplied to this incoming data also. Such a facility implies the presence of cognitive structures in the system. This means that the need for learning arises only when the incoming data is new and is not recognized as part of any existing rule. Learning occurs only on the differentials. Understanding thus arises on a mix of cognition and inference. Such understanding helps reduce long term processing and reprocessing data loads, which makes data matching and cognition by data analogy important processes in the path to understanding.

Understanding is thus both the generation of contextual or relational rules out of data and the perception of contexts from data and recall of its accompanying rule. Inference is learning, cognition is comprehension, the first part generates rules, and the second part uses it, such a two faced process results in reduced long term learning load. Natural entities in reality exhibit a mixed mode of understanding, part cognition and part inference.

Searle's Chinese Room argument (3) is basically about understanding, whether a formal computer program can be said to have understood intrinsically the meaning of the work it is doing. The inherent strength of Searle's argument is that it looks obvious to even the most casual observer of intelligence. Let us examine Searle's argument from our perspective.

Let us say that we have a learning system in an environment. If inferential processes are available on this system, then we can see that given data, then the system can deliver some results. (We assume such a system here, we know from our discussion in Part I, that such a natural system cannot originate or exist in nature)

Such a learning system is neither conscious nor intentional, it is more like a chemical reaction; a non-intentional non-conscious process that delivers a result that does not make sense either to the system or its environment. This kind of system can however be useful to an intentional intelligent entity that can understand what the system does and wishes to use its results. The user entity may, from an understanding of the inferential processes, modify the inputs or the processes to improve the systems functioning. If process improvements need to be done, then the intentional entity or the user has to treat the results of the system as feedback and provide corrective measures to tune the systems outputs to his requirements.

Here we see that a success monitoring system comes into operation. This success monitoring system is located in the user, who in effect becomes the master and acts on the slave system to correct it. On the whole we see that a directed learning process acts on the combination of master and slave to improve the system's results. Notice here that the user is intentional, the system isn't, however the master's intent enwraps the slave system. So for an external observer concentrating on the results, the user + system combo appears to be an intentional system. Both data cognition and result cognition rest with the master. Here we see that it is the master who understands both the data and the result of the slave system.

Can we remove the need for a master and make the system demonstrate implicit cognition? We did say that cognition implies that the system can look at an incoming data and match it to a known internal context or a context-based rule; it can either do data matching or rule matching. We see that the system is already equipped with inferential processes. These inferential processes can act on any data or on a limited set of data. For a start, we can provide some data boundaries to the system; such data that arrive within the bounds are only processed. Such a facility could help convince an external observer that there exists a primitive kind of default cognition in the machine.

However this is not true cognition, mere data straining. For true cognition, the system needs a data store to which incoming data can be matched. Since no data store exists or can arise out of default, it is clear that such a store and a data matching process should be provided to the system by an external agency or the user. Given such a data store and associated cognitive processes, the systems cognitive processes will act on all incoming data before they are passed on to its inferential processes. Here the system can demonstrate cognition, but this is a case of cognition transfer, from the master to the slave system. Here the cognitive load, which was earlier exclusively the domain of the master, is now shared between the master and the slave.

Can the system improve its inferential capacities by itself? From our earlier example we know that any improvement of learning or the associated inferential processes has to come from a directed learning process that in turn demands the presence of feedback and response matching mechanisms. As in the earlier case, it is possible for some feedback and auto correction mechanisms to exist in some systems by default. Such mechanisms even if they exist are more a gyroscopic correction rather than intentional correction. It is possible for a neural network like learning system to demonstrate improved learning, however learning improvements do not translate to improvements in inferential capacities. However it is to be agreed that the presence of feedback mechanisms and directed learning processes would enable the target system to move in a direction of increasingly maximizing its fit to a target or target environment.

Any real improvement in inferential capacities beyond the capacities already built in would need a superior process based user to transfer his inferential process knowledge to the system. Such a user driven improvement in inferential processes combined with appropriate feedback mechanisms can positively impact system performance. As in the cognition transfer case, here the inferential process loads are shared between the master and the slave.

A system that explicitly benefits from such cognitive and inferential process transfer also gains an unintended implicit benefit, the transfer of autonomy from master to slave; there arises a sharing of autonomy and control. Even here the system is not completely autonomous, it is autonomous with respect to its own domain.

We have till now considered three cases of intelligent systems, the first case where the system is non conscious and non intentional, the second case where the system is (partially) conscious but non intentional, and the third case where the system is (partially) conscious and intentional. Naturally the last case turns out to be (fully) conscious intentional systems.

Nature equips its systems with both consciousness and intent (of various levels depending on their organization), all natural systems fall in the last category, fully conscious and intentional. This combination is what makes these systems truly autonomous. This autonomy is not as in the earlier case confined to any particular domain within the target environment. Such autonomy is complete and does not suffer from any domain restrictions in a given environment.

Part 1 of this paper (1) discusses how life and the desire to sustain it may give rise to both consciousness and intent and how these artifacts are concomitant and connatural to all life and living systems. It discusses how consciousness and intent combine to create learning systems that can learn, act and improve themselves continuously to keep in pace with environmental and systemic challenges.

Consciousness and intent are so intertwined in natural systems that we fail to notice that they are actually two distinct artifacts. From our classification above, we see that it is possible for them to be separate and act separately. It is only with the rise of man made intelligent systems that we have been forced us to sit up and take notice of this distinction and consider its impact on intelligence systems and understanding.

Despite the lack of true autonomy the first three classes of systems can make for good artificial systems. Their dependence on human masters for performance improvements however relegates them to the role of slave systems. We however know that it is possible to build artificial autonomous conscious intentional systems at least in limited domains, which is in itself a major improvement.

For such an autonomous conscious intelligent system, (see Part 1) understanding arises out of dealing with and learning from contextually relevant data. This implies the presence of sensors that bring entity relevant data in, collation of such data into existing or newly created data stores, cognitive and inferential capabilities that are based on this sensor data/data store combination and an internal success monitoring system that aims to improve system responses. Such a system with a mix of cognitive and inferential capacities can continuously improve its responses until it hits learning limits.

Does this mean that we can build artificial intelligence systems that are equivalent to natural systems? The answer to this question, like the truth, lies somewhere in between. Given that the environment is changing and can even change drastically, our autonomous artificial system can successfully learn and adapt to the changing environment until it hits learning limits. Beyond these limits, the system cannot learn and if this system were to inhabit a natural environment, then it would fail in the face of increased learning demands.

Natural systems are also prone to this problem and when they fail en masse we term it extinction. Natural systems however have some facilities that the AI systems of today do not enjoy. Within learning limits and extinction limits, natural systems use the systems of reproduction and natural selection to produce entities that demonstrate continuously improving learning even with little processing power. However when these limits are hit, natural entities have the ability to modify/improve themselves and their systems to remain in tune with their environments. They seem to have the ability to seamlessly blend both these options in the drive to survive and flourish.

In other words these systems can do internal and external course correction. When environments change, even beyond the original conditions that these natural systems were designed for, then these systems have changed /improved/modified/adapted themselves to keep in step with the environment. Life imbues its tenants with a tenacious desire for survival that overarches all other demands, systemic and environmental.

It is this desire that forms the ground for all other activities of the system like consciousness, sustenance, reproduction, learning systems, intelligence, shape, size, etc. With this end point in view natural systems track themselves with respect to their environments and change course along with it, modifying their processes and procedures to remain in tune with it. Such a facility which does not exist for AI entities at present cannot however be ruled out in the future. However, this is only one side of the story.

The other side of the story, which is more important to us from the AI point of view, is that natural systems are not mere environment responsive entities. Rather than forever respond to their environments, many of them take charge of their local environments to the extent possible, modifying it and coercing it to remain within the comfort levels their systems can handle. The resultant reduction in environment variation directly translates to less learning loads and a more stress free life style. We have not begun to envisage such a course of action for our artificial entities; we stay the course of environment responsiveness and endeavor to build entities that are increasingly responsive to their environments.

Such self and environment modification behavior, which we term proactive behavior, allows the entity to handle a greater deal of environment variation than what its underlying intelligence systems are designed to handle. Proactive behavior (1) is the basis of the richness and the even seeming weirdness of most natural activity. In proactive behavior, the natural entity arm-twists the local environment to keep variations under its control; we humans are the more extreme examples of such behavior. Proactive behavior is what allows natural entities to display a modicum of comfort with their environments, what removes the need for perennially responsive behavior.

Natural intelligence systems thus come inbuilt with two advantages over artificial systems, one is the inbuilt facility for cognition and inference, and the other is the facility for course correction by natural selection. Proactive behavior acts in the interim to amplify the success of these mechanisms. Their other major advantage comes from reproduction and incubation, which enable archive based learning retention and help remove/reduce the starting loads that can stall any learning system introduced to a highly dynamic environment. With such an archive the entity can survive and flourish even in complex environments with very little dynamic online learning and thus little processing power.

How did such natural faculties and facilities arise in the first place? Unless one is prepared to assume out-of-thin-air theories or intelligent designer based theories, one is forced to depend on the theory of evolution as a logical explanation for the evolution of natural intelligence systems too. If the theory of evolution were correct, then one can imagine a logical learning system growth path that is co-evolutionary to evolution itself. Such a growth path envisages the growth of consciousness and intelligence in consonance with natural entity evolution from bacterial scales to human scales.

Our likely scenario posits generations of autonomous systems that developed sensors as they grew and inferential processes to make sense of these sensory data so that the learning could prove useful for sustenance and survival. Only such a natural evolutionary growth path aided by reproduction can result in the growth of implicit cognition and inference and also account for the variety of intelligence mechanisms on the planet.

Artificial systems that are not blessed with such inbuilt, implicit inference and cognitive growth paths will need to borrow the facilities of cognition and inference from naturally conscious and intelligent systems. This dependence of artificial systems on natural systems for improvement in inference and cognition means that they would remain slave systems. Despite the autonomy we can embed them with, their inability for course correction will ensure that they will either function well or go extinct. A facility for self-modification or systems improvement is a far way off for AI entities, what to say of natural selection. Prima facie, it may look as if these artificial systems are condemned to permanent slavery, but we will soon see that it need not always be so.

One nagging question that still remains is this; how is it possible that a system, however natural, however primitive, originate its own cognitive and inferential processes? One is forced to picture a start point. This naturally leads to a conjecture about the origins of life. However speculation about life's origins can take us too far away from topicality.

Overcoming Searle's Chinese Room forms a major stumbling block to the AI fantasy that one day we may be able to make human like systems. Human like systems should demonstrate human like understanding and Searle understandably finds some critical parts missing. Natural understanding, we did discuss hinges on the presence of both cognition and inference and our artificial entities miss parts of both, however the more yawning vacuum is in cognition. Searle's argument therefore should not have come as a surprise to AI enthusiasts.

In our discussion in Part 1 on conscious intelligent systems we discussed how the presence of consciousness forces the rise of learning systems and intelligence. We also discussed how the presence of life gives rise to intent. In fact life, consciousness, intent, intelligence and learning systems all seem to have a single thread run through them.

We can now see that in most of the artificial systems that we design, one of these two factors are lacking, either consciousness or intent. In reality, we would not find it difficult to locate most of our present day artificial entities in one of these three slave classes we discussed earlier. In theory one could transfer a human level of cognition and inference to a humanoid like entity, but our current understanding of our cognitive faculties is very much limited.

Our current understanding of our inference processes is better; there is a specific reason for such an asymmetric tilt, a point we discuss further down in this discussion. We have used this knowledge of our inference processes to build increasingly better slave systems. Notice that this knowledge of our inference processes tilts the design of our artificial entities towards symbol based discrete processing, while for natural entities there is little evidence that this is the case.

The truth is that we still do not know how human and other natural intelligence systems work at their most basic levels. However we know that symbol based discrete processing has inherent advantages that can help turn our artificial intelligence systems into better, faster and more relentless learners than humans, which is good reason to be wary of them. Deep Blue and such game based learning programs strengthen such a claim.

Humans, at least most of us, still face difficulties with symbol based processing, we are more comfortable with fuzzy heuristic learning and understanding. The final human advantage over the coming humanoid may rest on our more open architecture that is not really dependent on individual entity learning and our closer ties to the very environment that begot us. The question we need to discuss and will discuss in passing, right here in this document is, what does it take to be one. The results as you can see could be surprising.

**Communication of Understanding**

In his 1950 paper titled Computing Machinery and Intelligence, Turing (2) inherently recognizes the difficulties in matching human understanding to artificial understanding. Turing considered this difficult question in the same paper, where he noted the argument that to see the actual processes of a machine, one has to become the machine, the solipsist position he calls it, and he says that this is a difficult proposition to undertake; instead he suggests that we create a general classification of external behavior that can help determine what the machine is doing. This directly leads to the Interview Room approach.

The Turing paper is admittedly a little too lenient to the machine; it also makes a lot of implicit assumptions. For example Turing tacitly assumes that his systems would understand as well as man, if not in the same fashion as man. He tides over the difficult question of what actually constitutes thinking by avoiding what he calls a Gallup poll over it. This simplistic tiding over however introduces to the system to a lot of deficiencies, the most important of which is the one Searle pointed out, the possibility for the falsification of understanding. There is also an implied equivalence of learning to thinking, an equivalence that may not be true. Turing also tacitly assumes that the AI entity in the room is as autonomous as a human is, which we have seen is possible only for conscious, intentional, intelligence systems.

Searle's objective is to show that the Interview room approach is open to falsification. The Chinese room example shows that it is possible to demonstrate an activity that can be perceived by an external observer to have risen out of understanding even without any real understanding being done. Falsification by artificial entities is however not a new discovery; an interesting example is the story of the old Chinese humanoid automaton (4) that during its demonstration is said to have incensed the Chinese Emperor by winking at his queen and concubines. What is it between the Chinese and understanding?

In real life, one can express love without really being in love; similarly school classrooms and daily life provide ample demonstration of learning without understanding. The classroom is in many ways a Chinese room; students' deal with mostly non-contextually relevant topics and solutions and solution regurgitation or cramming provides an easy way to demonstrate understanding without understanding. Real life falsification of understanding is much more common than we would like to admit.

We did earlier claim that we could build a conscious intelligent system that can understand implicitly, say in the Searlian sense of the term. But this need not convince holders of the Searlian position; they may continue to hold that our system does not understand. Here in this portion we recognize that falsification of understanding can exist, we even say that real life Chinese Room scenarios are quite common and are taken for granted. How shall we prove that our system understands? How then can we distinguish between true understanding and false understanding? What is the real test of understanding?

Any test for understanding naturally implies two entities or situations, the tester and the tested. When two entities are present and one of them needs to test the others understanding, then the communication of understanding and its subsequent cognition/interpretation becomes the most important factor in understanding the level of understanding.

Ideally the communication of understanding should trigger both cognition of understanding and understanding in the recipient, but it need not always be so. There is a small gap inherent to the communication of understanding and the actual process of recipient understanding and falsification of understanding exploits this small gap.

To understand the presence of this gap, we have to consider the implications of the communication of understanding and the subsequent recipient understanding that should follow. We did see that for autonomous entities, understanding arises out of learning from contextual data. Such self-referential understanding we will call implicit understanding. All autonomous conscious intelligent entities, natural and artificial, therefore can and must demonstrate implicit understanding.

How can we, or another tester entity determine if a target entity has understood how to resolve a problem? The only method is by giving the tested entity a similar problem to solve. When it solves the given problem, we can say that the target entity has understood. This is all right as long as the verifying entity is intellectually superior to the verified entity. However in such a case the act of verification has little use other than verification.

The other side of the process is tester or recipient understanding. We can see that recipient understanding can occur only when the received data and solutions are retrofitted to the recipient's contextual database and the cognitive/inferential processes run on it to generate recipient inferences and understanding.

In real world scenarios, we look forward to asking others as to how a real world problem can be resolved. The demonstration of understanding can also be for teaching or training, because such training or teaching helps cut short the solution discovery process in the recipient by rule seeding. How will an entity that has understood something implicitly, communicate its knowledge to someone who needs it?

In the absence of a formal common language, and the absence of other formal communication channels, (like say telepathy or the Matrix), such communication can only be by physical demonstration of the solution within the limits of the sensor and output abilities. We recognize such physically oriented acts of learning as learning by imitation.

'Why,' said the Dodo, 'the best way to explain it is to do it.' (Lewis Carroll, Alice in Wonderland).

Imitation and learning by imitation is a key portion of communicating implicit understanding. In the absence of a formal communication artifact, implicit understanding can only be communicated by demonstration of similar problems and the solution path.

Does the conveyance and receipt of the solution path imply immediate understanding on the part of the recipient? It may or may not. Retrofitting the solution may even require acquisition of new data and its incorporation into the contextual templates; the process is necessarily difficult and time consuming.

In natural conditions the need for the solution may arise sooner than the time required for understanding. Even otherwise, natural intelligence systems, we did discuss in Part 1, are tuned for partial and iterative learning, which when extended means that much of our understanding could also be partial and will take time. Therefore, rather than go the whole way, it makes sense to carry the solution in memory and regurgitate it, even in the absence of proper understanding. In fact in natural organisms, this stimulus for imitation (and thus delayed learning) is strong and forms the core of most learning by training.

We can see that a buffer like facility for stored postponed learning and solution regurgitation can be a boon to most natural organisms stressed for time, survival and processing resources. Falsification may be a necessary artifact in a group of learning systems stressed for time and the resources required for faster understanding.

So why understand it when you can buffer it? Implicit understanding helps reduce the explicit memory carrying loads of such a buffer facility and helps release it for further data acquisition; even otherwise we did see that the main function of understanding is to help reduce data processing demands, w.r.t present and future data.

So how can the donor check if the recipient has truly understood his solution? By posing the problem back to the recipient and requesting an answer from it, not the same problem but a similar or related problem. If the problem is too similar and can be resolved by regurgitation, the recipient can still fool the tester or donor. Only when the problem is sufficiently difficult and the taught entity can use the solution path to relate to it and solve it can it be said to demonstrate true understanding.

Therefore we see that communication of understanding does not automatically imply understanding on the path of the recipient not only that the recipient can under some conditions simulate understanding by solution regurgitation, which means that the chance of falsification of understanding still exists.

The saving grace is that implicit communication has limitations that make it slightly difficult to feign understanding for long, the limitations of morphology and context. It has to be understood that for effective communication to happen between any two autonomous entities they need to share a contextual database. In nature, this commonality of the contextual database and morphology within a group is available and arises because of reproduction and speciation. Both processes ensure that all members of a group share similar contextual data, patterns, or pattern templates. Other limitations like environmental boundaries, sensory graininess, and morphological activity boundaries also act to constrain context spread. These limitations ensure that the scope of falsification is low in a group that can demonstrate only implicit understanding.

This will become clearer when we discuss the next type of understanding and its communication, which we term explicit. In explicit understanding, understanding is communicated not by imitation, but by using a formal symbol set, the human form of which we call language. When implicit understanding and its communication are effective, where does the need for explicit understanding arise? How does a formal symbol set emerge and why? Does the scope for falsification increase or decrease in the new case?

**Minds Necessitate Language**

Please see Part 1 of this paper (1) for a more detailed discussion on the reasons for the rise and presence of the mind and our sense of I. An outline of the salient points is given here because the rise of the mind affects understanding and its communication to a larger extent than what we usually consider.

The discussion in Part 1 equates functional consciousness to the presence of the sensory boundaries and argues that all natural intelligence systems gain a sense of basic consciousness through their sensors. It says that intelligence/learning systems arise to make rules and responses to/out of these sensory data. The repetitive nature of the environment forces the rise of archive based, pattern based learning systems because such systems aid long term processing efficiency. It says that natural environments with its urgent response demands tend to constrain dynamic online learning and force learning to be done offline during entity rest periods. This learning load shifting has given rise to a host of unintended benefits including the human presence on earth.

For offline learning to occur, the archived online data should be replayed to the learning system during rest time and the learning system learns from such regurgitated data. This replay of the data however results in the replay of the entity's sensory boundaries, which results in a recreation of the entity's sensor based consciousness itself. Though a shadow entity is generated in the system, the presence of this entity and the offline learning process allow the basic entity to offload its learning loads and concentrate on archive based environment response. The shadow entity and its offline processes however need proper identification and corralling that they do not interfere with the environmental responses of the main entity. Offline learning demands good rest and a considerable amount of entity safety. Higher the rest period and the safety, higher the learning potential of the entity and better its learning quality!

The maximization of learning potential occurs in the following scenario. First the entity finds a time of rest and safety so that it can shutdown its online response mechanisms and release processing power to the offline processing mechanisms. We equate this to sleep. Second, during the time it is awake, the entity shuttles between online and offline processing whenever it is at rest. We equate this condition to wakefulness.

# Conscious Intelligent Systems
Artificial Intelligence and Consciousness – A Learning System Perspective
## Part II: Mind, Thought, Language, and Understanding

We say that man's condition corresponds to such a maximized state. We also equate the offline learning theater to the mind and see the mind as a copy of simple consciousness. We equate the offline learning processes to mentation processes, the results of such processes as thoughts and the identity of the shadow entity to our sense of I. We discuss how proactive behavior tends to keep the mind busy. We use this above understanding to consider the effects that the rise of the mind has on understanding and its communication. For a more detailed explanation see Part 1(1).

In implicit understanding, the communication of understanding was done by imitation. Imitation in itself implies concurrency of the problem context and its solution. For imitation it is necessary that the problem and its solution be physically available and that the solution to a problem is demonstrated physically and learnt physically.

However with the rise of the mind, such a communication scenario becomes difficult if not impossible. The contexts, problems, and solutions of the mind generally refer to objects and problems that may not be physically available and are not current. For instance a problem encountered ten days back might have generated a solution at rest, in a circumstance where the artifacts and environment of the original problem are not only not present but also non re-creatable.

Also notice that since the learning process is mental and its results arise only as thoughts, it is invisible to other members of the group. These limitations in effect isolate understanding to each member of the group. The lack of communication of such mind-based understanding can be deleterious to a group that relies on social grouping to gain environmental benefits and ward off environmental dangers. On the other hand a facility for such communication can lead to a host of benefits.

More the presence of mind, more acute becomes the problem. Synchrony could be lost, individual activities can become suspect, the benefits that could be gained by cooperation are lost and misunderstanding can result. A method of communication of mental processes and its results becomes an extreme necessity for successful group coherence and activity.

It is obvious that before a communication process can arise, the identification of mental contexts and contents is necessary. It is also obvious that such identification should be common across the group to avoid chances of error and misunderstanding. It is logical that the identification should rely on a kind of mnemonic or shorthand to make the communication process short, clear, and efficient. It is obvious that such a kind of shorthand or symbolic naming can only arise across a group that shares a contextual database. It is however not essential that all these rise in sequence, they could rise in tandem too. Whatever be the sequence of steps, it is clear that these steps are necessary to create a communication system that can connect individuals equipped with minds.

We see that the presence of mind and a need for communication of mental contents necessitate the rise of language or language like communication artifacts. Such language like communication artifacts should be designed for speed, size, contextual spread, and clarity. Once such a language like ability rises, communication of mental artifacts and solutions become easy.

While language arises from the need for communication of mental contents, its effects are more far reaching than mere communication, we can say that it changes the very ground it arose from. Language and language like abilities tend to change the very nature of cognition and inference, the two pillars of understanding.

Notice that in implicit understanding, the processes of cognition and inference are essentially processes that the entity is blind to; it only reacts to the results and variables of the learning process, much as modern computer programs tend to do.

In implicit understanding, contexts are basically interconnected collections of data that tend to pile up, resulting in the general data arrangement being similar to a group of hills occupying an area. Processing is basically hill based and context driven. With the arrival of language, the hilltops that exist as an assortment of data and links acquire formal names and are recognized as such by their names.

# Conscious Intelligent Systems

Artificial Intelligence and Consciousness – A Learning System Perspective

## Part II: Mind, Thought, Language, and Understanding

An apple, which is in fact an assortment of sensory and event related data, acquires a name. Cognition is now easier and firmer but more importantly explicit. This makes processing easier for the offline mechanisms too, the mind, always under the pressure of being displaced by the basic consciousness process would find it easier to recognize things by their names or symbolic tags. The mnemonic nature of communication when combined with the need to communicate the solution and the solution path changes the nature of inference too. Inference also becomes explicit, an important point we will take up for discussion at a later stage.

We refer to this combination of explicit inference and cognition as explicit understanding. We see that the presence of the mind forces both understanding and the communication of understanding to the explicit foreground. The ever-present threat of roll back of the mind to make way for the online mechanisms force the mental processes to be run and described efficiently. This is probably one reason why thoughts rely on the presence and aid of language.

Thoughts, which we presume to be the interim and final results of offline processing, can perhaps act equally well on the contextual collections implicit to implicit understanding, but the demands of efficiency may have forced the rise of the use of language for the mental processes too. Therefore, it can be presumed that while thoughts may not need explicit language, they do better with language.

Here we see the cart lead the horse; the demand for communication of mental processes and the rise of the corresponding communication artifacts has affected the very learning process itself. One can presume that through the course of human evolution, the growing presence of the mind and the corresponding rise of language probably varied the very nature of the human learning process.

If our assumptions and the logic we base them on are correct, then we see that the mind, its thoughts, and language are as intimately connected as brains and minds. Minds necessitate language or language like symbolic communication artifacts. Can we claim that minds cause language?

The presence and complexity of language or language like formal symbol based communication abilities is a good indicator of the presence of mind and its level. This however means that all natural entities equipped with the offline learning mechanism are also possibly equipped with methods of formal symbol based language like communication. The range and complexity of such formal symbolism and language varies according to the presence of mind and the contextual spread the organism is privy to. This does imply that man is not the only natural entity that has the facility for language like formal symbol based communication, though he may be the best among the lot on account of the practically permanent presence of mind.

It can be presumed that on an evolutionary path, symbolic communication will initially tend to reuse the same output pathways that are used for implicit communication. However such output pathways could have limitations and may lead to the possibility of increased misunderstanding and error (dumb charades!). The chances for such error magnify when the contextual database explodes, as with say migration. The absence of an open communication pathways, independent of morphological peculiarities can act as a real dampener on group size, migration, and ultimate species success.

In such cases of limited communication capabilities, ideas spread over multiple contexts would be difficult to communicate, because each context will demand a particular configuration of signals. In the face of limited signaling capacities, this would mean an overload on existing signaling configurations. This would in turn mean that each signal could be interpreted in many ways; the most possible interpretation would however rest on the present context, as in poetry, confusion can and will result. Given the practically permanent presence of the mind, the advent of human vocalization seems to have let loose the contextual landscape that man can explore and seems to have aided the drive for migration and species success.

Now that we see the demand for communication of offline learning processes and minds as possible reasons for the rise of language, it is logical to ask what subjects or contexts would form the contents of language? From our perspective, we can see that the contents of any language like system are equivalent to the contents of the offline mentation processes. At a given point in time, the objects of these offline processes are the contexts/problems that occupy high positions on the learning stack; such contexts should necessarily have urgent functional or baroque targets. Higher the priority, more the time that will be spent by the offline processes as it strains to find a suitable solution. Higher the attention or priority on a particular context more fine-tuned and intricate will be the symbolic language artifacts attached to the object.

From such a perspective, we can see that most of the instinctual heritable stuff is not sweated on, after all solutions for such problems already exist in the inherited learning archive, unless the consciousness mechanism finds the existing solutions not good enough and decides to have a go at it. Wider the contextual spread of the organism, wider would be the spread of the explicit communication glossary. In an active language enabled community, one can expect that with time, an increasing number of artifacts get pushed to the explicit foreground and will thus acquire formal symbolic names. Language grows and enfolds most of the contexts and objects of the entity's activity.

Here we use the words implicit and explicit not in the sense of the antonym, but in the sense of extension. Offline learning processes occur as a result of learning load spill over, like magma flowing out of a volcano. As in magma flow, the explicit processes are not only a pointer to the internal processes, but make visible the internal processes, which would otherwise be invisible.

We did see that the presence of mind and language move cognition and inference from implicit to explicit. The effect of this motion on cognition and inference are not the same, they are quite different.

In the case of cognition, the explicit symbolic tag on a contextual hilltop effectively hides the contextual mass and interlinks that lies under the hilltop. An apple for instance may cognitively indicate the sum of a different collection of contexts for each individual, for some it may be A for Apple / Eve's apple / mom's apple pie / or raiding a neighborhood tree.

The symbolic naming process actually hides these details. This hiding and encapsulation, to use a C++ metaphor, works out to our disadvantage when programming for AI cognition, we really do not know how we recognize objects and therefore we find it difficult to map cognitive processes to our AI entities. The natural sciences are just beginning to get a grip on nature's cognitive processes.

On the other hand our knowledge of our inference processes has improved from this implicit to explicit motion. Perhaps this had more to do with explicit communication needs, where it was not necessary to describe what an apple was, (which everyone knows, in a given community, such cognition is taken for granted), however the process of stealing one needs quite some elaboration. This requirement for explicitness of the solution and the solution path seems to have improved our knowledge of how we actually infer. Did sitting and chatting round the campfire actually improve our explicit knowledge of inferential processes?

Over time and experience we have learnt more about our inferential processes and have almost standardized much of them that are linear and predictable. It is out of this knowledge of the inference processes that math, science, and other subjects have evolved. It is exactly this knowledge that we have used to build slave systems of increasing complexity and intelligence. It is this knowledge that we seek to embed in our AI entities. Thus, it is the rise of the mind and more importantly its exposure of our inferential processes that have combined to carry us far away from our animal roots into realms that none of nature's other intelligence systems have been able to venture into. We also see why they are not able to or would be able to.

The author suspects that minds and language are not unique to man, other entities also show signs of having them; a good indicator would perhaps be the presence of sleep. These entities would be able to demonstrate a mix of explicit and implicit understanding, if but primitive. Notice that humans as a species are still not comfortable with explicit understanding; explicit inference creates more problems for us than any, so much that our computers tend to do it better and faster. However we do quite well on the explicit cognition scale, better than most animals.

If we mark a scale with implicit understanding on one end and explicit understanding on the other, then we can see that humans would fare no better than slightly above the mid level mark. There will also be a large gap between us humans and our immediate existing predecessors, the advanced primates, the intervening area probably being filled in by our extinct predecessors. It is the presence of this large gap that sparks in us a distinct sense of separateness from our predecessors.

Which is the better learning path, implicit or explicit? The answer cannot be very clear-cut or straightforward. We can see that during wakefulness there is a permanent pressure on the presence of mind; the threat of recall is always there. There is also an ever-present pressure on the processing resources that can be parceled out to an offline process. This pressure, when in the presence of symbolic artifacts can lead to abstraction and idealization of the knowledge frame. In scaled learning (1) terms we are forced to hover over the hilltops.

Only when we sleep does the competition for learning resources subside, in such a condition the need for explicit identification of the offline process is also reduced to a minimum, our sense of I can go away, the mental processes can be unhooked from the demand for abstraction and we can travel down the hill slopes. Whether this unhooking and pressure release does lead to a better solution is still a moot point. We will understand this process better if we alter the question to ask if intuition, which is the result of an unseen underground learning process is better than the explicit solution seeking process. The best answer could be, well it depends.

Oops, did we forget falsification while we were discovering the roots of language? We did see that there was a gap between sender communication and recipient understanding and how imitative facilities may allow us to use the solution without understanding it. We also did discuss that in implicit communication and understanding, the gap is generally small due to the limitations of morphology and context.

In explicit communication with its multiplicity of contexts, the gap widens and the necessity for an imitative solution storage facility increases. We did say that falsification, which may even be a necessity, exploits this gap between solution communication and understanding. We can therefore say that the scope of falsification rises with explicit communication and understanding, particularly when the contextual spread rises. And as in the solution for a check for implicit understanding, the only way to test it is to give the tested a problem of increasing difficulty in the context in which it is tested.

On the whole, the simple fact that only testing can help determine the level of understanding tells us that there is no avoiding the Interview Room approach, Turing was right in choosing it. The only way to reduce the chances of falsification is to use graded and more importantly context based testing, like we do in the real world. Once we understand that all understanding processes are actually partial and depend on the depth of the enquiry, this falsification may be seen as unavoidable. The level of falsification will then be inversely proportional to the level of understanding and vice versa. It is clear that a better understanding of understanding will help light up the path.

**The Humanoid Fantasies**

Most people who have been exposed to humanoids in stories and movies tend to implicitly assume that machines can speak like humans do. That these humanoids can also think is implied by assumption. We tend to take thought and speech for granted; even theorists in the field of AI cannot be absolved of making such anthropocentric assumptions. Most of us tend to equate thinking with learning and that partly accounts for the incredulity of AI practitioners who are baffled by people who claim that machines cannot think.

Now that we see that the presence of mind give rise to thoughts and language, we should know that our learning AI entities, however good learners they be, are not really thinking. And in the absence of thought they could not be speaking either! Not in human fashion anyway!

# Conscious Intelligent Systems
Artificial Intelligence and Consciousness – A Learning System Perspective
## Part II: Mind, Thought, Language, and Understanding

To have thoughts and language they need minds and we see that minds arise out of an entirely different set of constraints and enablers than what we see for our usual AI machines. Searle did point this out; the very fact that most AI theorists do not see it as obvious makes Searle's question look almost like a revelation. We can however see that his point is valid and very important to high-level AI design! That is if we plan to make humanoids!

Of the many implicit assumptions that Turing makes on his "Can Machines Think" question, the lack of thought of the very process of thought and its origins is glaring. In this 1950 paper, his entities think and speak by presumption, though there are no grounds to suppose they could think or speak.

If our hypothesis is correct, then we see that minds arise as a result of enfolded offline learning processes. These enfolded offline learning processes arise as a result of environmental constraints that deny online learning systems good learning time and opportunity. Thoughts and language arise out of the pressures of communication between such mind-enabled entities.

If the author's understanding is correct, there are no AI entities presently available or under construction that are embedded with such enfolded learning processes or mind based communication needs. Therefore with due respect to Turing and the others who are in agreement with him, the right answer to the question as to whether machines can think (in human fashion) is still a big NO. It naturally follows that they cannot speak too, in human fashion that is. Please read them sentences carefully, the inhabitants of Chinese and Interview rooms are yet to be born.

So what about those immensely capable machines that we have built and used. These are learning machines at best, like natural learning systems, artificial learning systems can even learn without thinking. However complex the learning process, it does not correspond to thought. Human thought and language arise out of a different design path. This is a critical difference and with a little thought you can see that the impact of this difference runs very deep.

Present day AI entities do not have minds and cannot speak, nor is there really reason to. Their communication between themselves and with us can at best be only implicit. However when they acquire minds, thoughts and language and need to communicate them, there is little guarantee that we will be able to understand them. To do all that however, we see that they should be able to make explicit their own mentation processes and relate to them. Now this automatically brings us to the question as to who will see these mentation processes.

The question of self-consciousness as you can see is quite a persistent one. Not only does it hold for us humans and trouble us so, it also extends to our yet to be born humanoids. We have tried to resolve part of this problem in Part 1 of this discussion. We can see that to speak and think like humans, AI entities also need the kind of self-sentience that humans have.

Such self-sentience first requires the presence of a binary consciousness (1). In the presence of binary consciousness, the sense of self-consciousness arrives, not out of thin air, but by constant correlation between the mental processes and the everyday acts and problems faced by the entity itself. Our children learn to do this early in life, they learn to correlate between their mental processes and their bodies and thus acquire self-consciousness. The evolutionary processes themselves must have long laid the base for such an easy acquisition during childhood; a long process we presume must have taxed our hominid ancestors so. AI entities will need to follow a similar acquisition pathway if we need them to speak and think like we do.

It is obvious that such a process, once its roots are understood, is not very difficult to implement in AI entities. The effects of self-sentience in humans on the world at large need consideration before we attempt such an implementation. Self-sentience in smaller AI entities may be beneficial to humans in the short term, but considering human history, the long-term effects may not be something that we would like to bet our species on.

Searle did propose that natural understanding is different from artificial understanding and that artificial systems may never understand as humans do. From our perspective we see three problems. One is the lack of cognitive ability, another is the lack of mind, the third is however more important, the ability to watch the environment and reset its goals, programs and systems to the changing environment.

We have seen that the cognitive gap can be closed, if not now then at least later when our knowledge of our own cognitive processes improves. The second problem, the rise of the mind is what we have just discussed. The third is a tougher option, to watch the changing environment and tag along with it, changing the systems goals and program paths and perhaps components and systems. It requires a blind persistence that life exhibits, to survive whatever, wherever, however. That would be the last frontier.